\documentclass[letterpaper]{article}
\usepackage{uai2019}
\usepackage[margin=1in]{geometry}
\usepackage{times}
\usepackage{natbib}
\usepackage{graphicx}
\usepackage{hyperref}
\usepackage[dvipsnames]{xcolor}
\usepackage{url}
\usepackage{amsfonts,amssymb,amsthm,amsmath}
\usepackage{bm}
\usepackage[inline]{enumitem}
\usepackage{tikz}
\usepackage{algorithm}
\usepackage[noend]{algpseudocode}
\usepackage{xpatch}

\makeatletter
\xpatchcmd{\algorithmic}{\itemsep\z@}{\itemsep=0.5ex plus1pt}{}{}
\makeatother

\usetikzlibrary{cd,arrows,calc,shapes,positioning}
\tikzstyle{obs} = [circle,fill=white,draw=black,inner sep=1pt,minimum size=20pt,font=\fontsize{10}{10}\selectfont,node distance=1,thick]
\tikzstyle{latent} = [obs,dotted]
\tikzstyle{protected} = [obs,text=Orange,draw=Orange]
\tikzstyle{unfair} = [obs,text=BrickRed,draw=BrickRed]
\tikzstyle{target} = [obs,text=MidnightBlue,draw=MidnightBlue]
\tikzstyle{feature} = [obs,text=ForestGreen,draw=ForestGreen]

\newcommand{\edge}[3][]{ %
  \foreach \x in {#2} { %
    \foreach \y in {#3} { %
      \path (\x) edge [->, >={triangle 45}, #1,thick] (\y) ;%
    } ;
  } ;
}

\newcommand{\bR}{\mathbb{R}}
\newcommand{\cG}{\mathcal{G}}
\newcommand{\cV}{\mathcal{V}}
\newcommand{\cE}{\mathcal{E}}

\DeclareMathOperator*{\argmin}{arg\,min}
\DeclareMathOperator*{\diag}{diag}

\DeclareMathOperator{\indep}{\perp\!\!\!\perp}
\DeclareMathOperator{\dep}{\not\! \perp\!\!\!\perp}

\newcommand{\B}[1]{\bm{#1}}

\DeclareMathOperator{\CFU}{\ensuremath \mathrm{CFU}}

\newcommand{\pa}{\ensuremath \mathrm{pa}}
\newcommand{\cD}{\ensuremath \mathcal{D}}

\title{The Sensitivity of Counterfactual Fairness to Unmeasured Confounding}

\author{%
{\bf Niki Kilbertus} \\
MPI for Intelligent Systems \\
University of Cambridge \\
\And
{\bf Philip~J.~Ball} \\
University of Cambridge \\
Microsoft Research \\
\And
{\bf Matt~J.~Kusner} \\
University of Oxford \\
The Alan Turing Institute \\
\AND
{\bf Adrian Weller} \\
University of Cambridge \\
The Alan Turing Institute \\
\And
{\bf Ricardo Silva} \\
University College London \\
The Alan Turing Institute
}

\begin{document}

\maketitle

\begin{abstract}
\noindent
Causal approaches to fairness have seen substantial recent interest, both from the machine learning community and from wider parties interested in ethical prediction algorithms.
In no small part, this has been due to the fact that causal models allow one to simultaneously leverage data and expert knowledge to remove discriminatory effects from predictions.
However, one of the primary assumptions in causal modeling is that you know the causal graph.
This introduces a new opportunity for bias, caused by misspecifying the causal model.
One common way for misspecification to occur is via \emph{unmeasured confounding}: the true causal effect between variables is partially described by unobserved quantities.
In this work we design tools to assess the sensitivity of fairness measures to this confounding for the popular class of non-linear additive noise models (ANMs).
Specifically, we give a procedure for computing the maximum difference between two counterfactually fair predictors, where one has become biased due to confounding.
For the case of bivariate confounding our technique can be swiftly computed via a sequence of closed-form updates.
For multivariate confounding we give an algorithm that can be efficiently solved via automatic differentiation.
We demonstrate our new sensitivity analysis tools in real-world fairness scenarios to assess the bias arising from confounding.

\end{abstract}

\section{INTRODUCTION}
\label{sec:intro}

Most work on fairness in machine learning focuses on discrimination against subpopulations in high-stakes decisions such as in criminal justice, lending, and insurance~\citep{kamiran2009classifying,kamishima2012fairness,hardt2016equality,zafar2017fairness2,Berk2017}.
These subpopulations are defined by one or more \emph{protected attributes} such as race, gender, age, and sexual orientation.
More recently, causal reasoning has been introduced as a valuable tool for the detection and mitigation of harmful biases and disparities in machine learning systems.
Notably, it helped refine our understanding of two particular issues.

First, it has been shown by \citet{kleinberg2016inherent} and \citet{chouldechova2017fair} that various subsets of popular observation-based parity notions, which are based only on the joint distribution of all variables involved, can only be satisfied simultaneously in unrealistically trivial situations.
This leaves us choosing among criteria when all of them seem desirable.
However, since two different data generation mechanisms can give rise to the same observed joint distribution, observation-based notions cannot distinguish scenarios that may have very different fairness interpretations \citep{hardt2016equality}.

Second, an earlier approach to individual fairness by \citet{dwork2012fairness} based on the appealing postulate that ``similar individuals should be treated similarly'' has proven hard to operationalize in practice.
Specifically, it shifts the issue from defining what is fair to defining similarity with respect to the task at hand both between individuals as well as between outcomes.

By (a) explicitly modelling the underlying data generating mechanism with causal models, and (b) using causal primitives such as interventions and counterfactuals to formalize ``similar individuals'', causality provides valuable insights to resolve these conceptual roadblocks \citep{kilbertus2017avoiding,kusner:17,nabi2017fair,zhang2018fairness}.
In this work, we will focus on \emph{counterfactual fairness} as introduced by \citet{kusner:17}, an individual-specific criterion aimed at answering the counterfactual question: ``What would have been my prediction if---all else held causally equal---I was a member of another protected group?''.
Despite the utility of such causal criteria, they are often contested, because they are based on strong assumptions that are hard to verify in practice.
First and foremost, all causal fairness criteria proposed in the literature assume that the causal structure of the problem is known.
Typically, one relies on domain experts and methods for causal discovery from data to construct a plausible causal graph.
While it is often possible with few variables to get the causal graph approximately right, one often needs untestable assumptions to construct the full graph.
The most common untestable assumption is that there is no unmeasured confounding between some variables in the causal graph.
Because we cannot measure it, this confounding can introduce bias that is unaccounted for by causal fairness criteria.

In this work we propose a solution.
We introduce tools to measure the sensitivity of the popular \emph{counterfactual fairness} criterion to unmeasured confounding.
Our tools are designed for the commonly used class of non-linear additive noise models \citep[ANMs,][]{hoyer2009nonlinear}.
Specifically, they describe how counterfactual fairness changes under a given amount of confounding.
The core ideas here described can be adapted for sensitivity analysis of other measures of causal effect, such as the average treatment effect (ATE), itself a topic not commonly approached in the context of graphical causal models.
A discussion will be left for a future journal version of this paper.
Note that counterfactual fairness poses extra challenges compared to the ATE, as it requires the computation of counterfactuals in the sense of \cite{pearl:00}.
Concretely, our contributions are:
\begin{itemize}
  \item For confounding between two variables, we design a fast procedure for estimating the worst-case change in counterfactual fairness due to confounding.
  It consists of a series of closed-form updates assuming linear models with non-linear basis functions.
  This family of models is particularly useful in graphical causal models where any given node has only few parents.
  \item For more than two variables, we fashion an efficient procedure that leverages automatic differentiation to estimate worst-case counterfactual fairness.
  In particular, compared to standard  sensitivity analysis \citep[typically applied to ATE problems, see e.g.][]{dorie2016flexible}, we formulate the problem in a multivariate setting as opposed to the typical bivariate case.
  The presence of other modeling constraints brings new challenges not found in the standard literature.
  \item We demonstrate that our method allows us to understand how fairness guarantees degrade based on different confounding levels.
  We also show that even under high levels of confounding, learning counterfactually fair predictors has lower fairness degradation than standard predictors using all features or using all features save for the protected attributes.\footnote{Code to reproduce the results can be found at \href{https://github.com/nikikilbertus/cf-fairness-sensitivity}{github.com/nikikilbertus/cf-fairness-sensitivity}.}
\end{itemize}

\section{BACKGROUND}
\label{sec:background}

\subsection{CAUSALITY AND FAIRNESS}

We begin by describing key background in causal inference and reviewing the notion of counterfactual fairness \citep{kusner:17}.

\paragraph{Causality.}
We will use the \emph{structural causal model} (SCM) framework described in \citet{pearl:00}, and look at a popular subclass of these models called \emph{additive noise models} (ANMs) \citep{hoyer2009nonlinear}.
Specifically, an SCM is a directed acyclic graph (DAG) $\cG = (\cV,\cE)$ model, with nodes $\cV$ and edges $\cE$.
Each node $X \in \cV$ is a random variable that is a non-linear function $f_X$ of its direct parent nodes $\mbox{pa}_{\cG}(X)$ in $\cG$, plus additive error (noise) $\epsilon$ as follows: $X = f_X(\mbox{pa}_{\cG}(X)) + \epsilon$.
To make model fitting efficient we will consider (a) functions $f_X$ that derive all their non-linearity from an embedding function $\B{\phi}$ of their direct parents, and are linear in this embedding;
and (b) Gaussian error (noise) $\epsilon$ so that:
\begin{align}
X = \B{\phi}(\mbox{pa}_{\cG}(X))^{\top} \B{w}_X + \epsilon, \;\;\;s.t.\;\;\; \epsilon \sim \mathcal{N}(0, \sigma_X), \nonumber
\end{align}
where $\B{w}_X$ are weights.
Later on, we will consider ANMs over observed variables, where the errors may be correlated.
Note that this class of ANMs is not closed under marginalization.
For a more detailed analysis of the testable implications of the ANM assumption, see \citep{Peters2017}.
Neither of our choices (a) and (b) are a fundamental limitation of our framework: the framework can easily be extended to general non-linear, or even non-parametric functions $f_X$, as well as non-Gaussian errors.
In this work, we make this choice to balance flexibility and computational cost.

\paragraph{Counterfactual fairness.}
A recent definition of predictive fairness is counterfactual fairness (CF) \citep{kusner:17}, which to facilitate exposition focuses on total effects.
See \citep{nabi2017fair,chiappa:18} for an exploration of path-specific effects.
Intuitively, CF states that a predictor $\hat{Y}$ of some target outcome $Y$ gives you a fair prediction if, given that you are a member of group $a$ (i.e., race, gender, sexual orientation), it would have given you the same prediction (in probability) had you been a member of a different group $a'$.
This is formalized via the causal notion of \emph{counterfactuals} as follows:
\begin{align}
    P(\hat{Y}_{A\leftarrow a'} = y \;&|\; X= \B{x}, A=a) = \label{eq:cf} \\ P(\hat{Y}_{A\leftarrow a} = y \;&|\; X=\B{x}, A=a), \nonumber
\end{align}
where $\hat{Y}_{A\leftarrow a'}$ is the counterfactual prediction, imagining $A\!=\!a'$ (note that, because in reality $A\!=\!a$, we have that $\hat{Y}_{A\leftarrow a}\!=\!\hat{Y}$), and $\B{x}$ is a realization of other variables in the causal system.
In ANMs $\hat{Y}_{A\leftarrow a'}$ can be computed in four steps:
1. Fit the parameters of the causal model using the observed data: $\mathcal{D}\!=\!\{\B{x}_i,a_i\}_{i=1}^n$;
2. Using the fitted model and data $\mathcal{D}$, compute all error variables $\B{\epsilon}$;
3. Replace $A$ with counterfactual value $a'$ in all causal model equations;
4. Using parameters, error variables, and $a'$, recompute all variables affected (directly or indirectly) by $A$, and recompute the prediction $\hat{Y}$.
To learn a CF predictor satisfying eq.~\eqref{eq:cf} it is sufficient to use any variables that are non-descendants of $A$, such as the error variables $\B{\epsilon}$ \citep{kusner:17}.
\citet{loftus2018causal} provide further arguments for the importance of causality in fairness as well as a review of existing methods.

\paragraph{Unmeasured confounding.}
One key assumption on which CF relies is that there is no \emph{unmeasured confounding} relationship missing in the causal model.
In this work, we formalize unmeasured confounding as non-zero correlations between any two error variables in $\B{\epsilon}$ which are assumed to follow a multivariate Gaussian distribution.
Without accounting for this, the above counterfactual procedure will compute error variables that are not guaranteed to be independent of $A$.
Thus any predictor trained on these exogenous variables is not guaranteed to satisfy counterfactual fairness eq.~\eqref{eq:cf}.
This setup captures the idea that often we have a decent understanding of the causal structure, but might overlook confounding effects, here in the form of pairwise correlations of noise variables.
At the same time, such confounding is often unidentifiable (save for specific parameterizations).
Thus assessing confounding is not a model selection problem but a sensitivity analysis problem.
To perform such analysis we propose tools to measure the worst-case deviation in CF due to unmeasured confounding.
Before describing these tools, we first place them in the context of the long tradition of sensitivity analysis in causal modeling.

\subsection{TRADITIONAL SENSITIVITY ANALYSIS}

Sensitivity analysis, for quantities such as the average treatment effect, can be traced back at least to the work by Jerome Cornfield on the General Surgeon study concerning the smoking and lung cancer link \citep{rosenbaum:02}.
Rosenbaum cast the problem in a more explicit statistical framework, addressing the question on how the ATE would vary if some degree of association between a treatment and a outcome was due to unmeasured confounding.
The logic of sensitivity analysis can be described in a simplified way as follows:
i) choose a level of ``strength'' for the contribution of a latent variable to the structural equation(s) of the treatment and/or outcome;
ii) by fixing this confounder contribution, estimate the corresponding ATE;
iii) vary steps i) and ii) through a range of ``confounding effects'' to report the level of unmeasured confounding required to make the estimate ATE be statistically indistinguishable from zero;
iv) consult an expert to declare whether the level of confounding required for that to happen is too strong to be plausible, and if so conclude that the effect is real to the best of one's knowledge.
This basic idea has led to a large literature, see \citep{dorie2016flexible,Robins2000} as examples among many of the existing state-of-the-art papers on this topic.

Note the crucial difference between sensitivity analysis and just fitting a latent variable model: we are not learning a latent variable distribution, as the confounding effect for a single cause-effect pair is \emph{unidentifiable}.
By holding the contribution of the confounder as constant and known, the remaining parameters become identifiable.
We can vary the sensitivity parameter without assuming a probability measure on the confounding effect.
The hypothesis test mentioned in the example above can be substituted by other criteria of practical significance.

Much of the work in the statistics literature on sensitivity analysis addresses pairs of cause-effects as opposed to a causal system with intermediate outcomes, and focuses on the binary question on when an effect is non-zero.
The \emph{grid search} idea of attempting different levels of the confounding level does not necessarily translate well to a full SCM: grid search grows exponentially with the number of pairs of variables.
In our problem formulation described in the sequel, we are interested in bounding the maximum magnitude $p_{\max}$ of the error correlation matrix entries, while maximizing a measure of counterfactual unfairness to understand how it varies by the presence of unmeasured confounding.
The solution is not always to set all entries to $p_{\max}$, since among other things we may be interested in keeping a subset of error correlations to be zero.
In this case, a sparse correlation matrix with all off-diagonal values set to either 0 or $p_{\max}$ is not necessarily positive-definite.
A multidimensional search for the entries of the confounding correlation matrix is then necessary, which we will do in Section \ref{sec:multivariate} by encoding everything as a fully differentiable and unconstrained optimization problem.

\section{TOOL \#1: GRID-BASED}
\label{sec:bivariate}

The notion of sensitivity analysis in a SCM can be complex, particularly when the estimated quantity involves counterfactuals.
In this section, we first describe a tool that estimates the effect of confounding on counterfactual fairness, when the confounding is limited to two variables (i.e., \emph{bivariate confounding}).
This procedure is computationally efficient for this setting.
For the general setting of confounding between any number of variables (\emph{multivariate confounding}) we will introduce a separate tool in Section~\ref{sec:multivariate}.
Below we describe our fast two-variable tool using a real-world example.

\subsection{A MOTIVATING EXAMPLE}

To motivate our approach, let us revisit the example about law school success analyzed by \citet{kusner:17}.
In this task, we want to predict the first year average grade ($Y$) of incoming law school students from their grade-point average ($G$) before entering law school and their law school admission test scores ($L$).
In the original work, the goal was to train a predictor $\hat{Y}$ that was counterfactually fair with respect to race.

\begin{figure}
    \flushright
    \begin{tikzpicture}
    \node[protected, label=above:{\color{Orange}protected}] (A) {$A$};
    \node[latent, left=of A] (EG) {$\epsilon_G$};
    \node[latent, right=of A] (EL) {$\epsilon_L$};
    \node[feature, below left=0.5 of A] (G) {$G$};
    \node[feature, right=of G, label=right:{\color{ForestGreen}features}] (L) {$L$};
  \edge {A} {G,L};
    \edge {G,EL} {L};
    \edge {EG} {G};
    \node[left=of G, align=center] (label) {\textbf{Model A} {\color{black!75}(guessed)}\\[3pt]
    $\epsilon_G \indep \epsilon_L$ with\\[3pt]
        $\epsilon_G \sim \mathcal{N}(0,\sigma_G^2)$,\\ $\epsilon_L \sim \mathcal{N}(0,\sigma_L^2)$\\%
    };
    \end{tikzpicture}\\
    \vspace{-0.3cm}
    \begin{tikzpicture}[thick]
    \node[protected] (A) {$A$};
    \node[latent, left=of A] (EG) {$\epsilon_G$};
    \node[latent, right=of A] (EL) {$\epsilon_L$};
    \node[feature, below left=0.5 of A] (G) {$G$};
    \node[feature, right=of G, label=right:{\color{white}features}] (L) {$L$};
  \edge {A} {G,L};
    \edge {G,EL} {L};
    \edge {EG} {G};
   \draw[dashed, <->] (EG) to[bend left] (EL);
    \node[left=0.7 of G, align=center] (label) {\textbf{Model B} {\color{black!75}(true)}\\[3pt]
    $\epsilon_G \dep \epsilon_L$ with\\[3pt]
    $(\epsilon_G, \epsilon_L)^{\top} \sim \mathcal{N}(\B{0}, \Sigma)$\\\mbox{ }\\%
    };
    \end{tikzpicture}
    \caption{Causal models for the law school example.
    Model A is the guessed model that has no unobserved confounding.
    Model B includes confounding via the covariance matrix $\Sigma$, which is captured by a bidirected edge using the standard acyclic directed mixed graph notation \citep[ADMG,][]{richardson:03}.
    Our techniques will estimate the worst case difference in the estimation of counterfactual fairness due to such confounding (we will consider a more complicated setup in Section~\ref{sec:experiments}).}
    \label{fig:bivariate}
\end{figure}
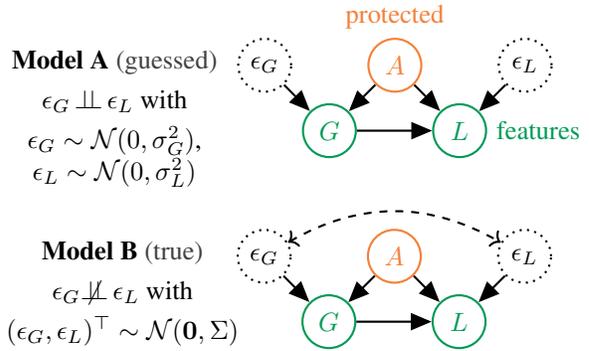

To evaluate any causal notion of fairness, we need to first specify the causal graph.
Here we assume $G \to L$ with errors $\epsilon_G, \epsilon_L$, where $G$ and $L$ are both influenced by the sensitive attribute $A$, see \textbf{Model A} in Figure~\ref{fig:bivariate}.
Given this specification, the standard way to train a counterfactually fair classifier is using $\epsilon_G, \epsilon_L$---the non-descendants of $A$.
To do so, we first learn them from data as the residuals in predicting $G$ and $L$ from their parents.

The validity of causal estimates rely on the assumption that the constructed causal model and its respective graph (here Model A) captures the true data-generating mechanism.
While previous work addressed how to enforce counterfactual fairness across a small enumeration of identifiable competing models \citep{russell:17}, in this work we consider misspecification in the lack of \emph{unidentifiable} unmeasured confounding.
In our example, this means violation of the assumed independence of the error variables $\epsilon_G$ and $\epsilon_L$.

To capture such confounding, we introduce \textbf{Model B} in Figure \ref{fig:bivariate}.
Here the error variables are not independent, they co-vary: $(\epsilon_G, \epsilon_L)^{\top}\!\sim\!\mathcal{N}(\B{0}, \Sigma)$ where,
\begin{equation*}
    \Sigma =
    \begin{pmatrix}
        \sigma_G^2 & p\, \sigma_G\, \sigma_L \\
        p\, \sigma_G \, \sigma_L & \sigma_L^2
    \end{pmatrix} .
\end{equation*}
Here, $\sigma_{\bullet}$ is the standard deviation of $\bullet$ and $p \in [-1, 1]$ is the correlation, such that the overall covariance matrix $\Sigma$ is positive semi-definite.
Before going into the detailed procedure of our sensitivity analysis, let us give a general description of what we mean by Model A and Model B throughout this work.

\textbf{Model A} is the ``guessed'' causal graph model used to build a counterfactually fair predictor.

\textbf{Model B} is a version of Model A that allows for further unobserved confounding between pairs of error variables not originally featured in A.
Model B will play the role of a hypothetical ground truth that simulates ``true'' counterfactual versions of the predictions based on Model A.

Our tool allows us to answer the following question: how does a predictor that is counterfactually fair under Model A perform in terms of counterfactual unfairness under the confounded Model B?
Our goal is to quantify how sensitive counterfactual unfairness is to misspecifications of the causal model, in particular to unobserved confounding.
To do so, we will introduce a measure which we will call \emph{counterfactual unfairness} (CFU).
Given this, we describe how to compute the worst-case violation of counterfactual fairness within a certain confounding budget, which we characterize by the correlation $-1 \leq p_{\max} \leq 1$ in Model B.
By varying the confounding budget, we can assess how robust Model A is to different degrees of model misspecification.
Like in classical sensitivity analysis, we can alternatively start from a level of unacceptable CFU, search for the minimum $p_{\max}$ whose worst-case CFU reaches this level, and leave it to domain experts to judge the plausibility of such a degree of unmeasured confounding $p_{\max}$.

\subsection{NOTATION AND PROBLEM SETUP}

For both Model A and B the model equations are:
\begin{equation}\label{eq:law}
    G = \B{\phi}_G(A)^{\top} \B{w}_{G} + \epsilon_G,\;\;
    L = \B{\phi}_L(A,G)^{\top} \B{w}_{L} + \epsilon_L ,
\end{equation}
where $\B{\phi}_G: \mathcal{A} \to \bR^{d_G}$ and $\B{\phi}_L: \mathcal{A} \times \bR \to \bR^{d_L}$ denotes \emph{fixed} embedding functions for $A$ and $A, G$ respectively, $A \in \mathcal{A}$ indicates the membership in a protected group (where $\mathcal{A}$ is the set of possible groups), and $\B{w}_{G} \in \bR^{d_G}$, $\B{w}_{L} \in \bR^{d_L}$ are the weights of the model.

In order to simplify notation, for observed data $\{(a_i, g_i, l_i)\}_{i=1}^n$, we define
\begin{align}\label{eq:notationbivariate}
\begin{split}
    \B{x}_i &=
    \begin{pmatrix}
        g_i\\
        l_i
    \end{pmatrix} \in \bR^{2}
    ,\quad
    \B{w} =
    \begin{pmatrix}
        \B{w}_{G} \\
        \B{w}_{L}
    \end{pmatrix} \in \bR^{d_G + d_L}
    ,\\
    \Phi_i &=
    \begin{pmatrix}
        \B{\phi}_{G_i}^{\top} &\B{0}^{\top} \\
        \B{0}^{\top} & \B{\phi}_{L_i}^{\top}
    \end{pmatrix} \in \bR^{2 \times (d_G + d_L)} ,
\end{split}
\end{align}
where we write $\B{\phi}_{G_i} = \B{\phi}_G(a_i)$ and $\B{\phi}_{L_i} = \B{\phi}_L(a_i, g_i)$ for brevity.
In eq.~\eqref{eq:notationbivariate} as well as the remainder of this work, equations and assignments with subscripts $i$ on both sides hold for all $i \in \{1, \ldots, n\}$.\footnote{Note that $A$ need not be exogenous.
Since we would need to include additional---standard but occluding---steps in the algorithm to handle discrete variables, this assumption is solely to simplify the presentation.}

\subsection{MODEL A: FIT CF PREDICTOR}
\label{subsec:model_a}

First, we build a counterfactually fair predictor with our guessed unconfounded Model A via the following steps.
\begin{enumerate}[label=\textbf{\arabic*.},wide,labelwidth=!,labelindent=0pt]
    \item Fit Model A via regularized maximum likelihood:
    \begin{align}\label{eq:modela_optimization}
    \begin{split}
        \min_{\B{w}, \sigma_G, \sigma_L} &\sum_{i=1}^n (\B{x}_i - \Phi_i \B{w})^{\top} \Sigma^{-1} (\B{x}_i - \Phi_i \B{w}) \\
        &+ \lambda \|\B{w}\|_2^2 + n \log \det (\Sigma) ,
    \end{split}
    \end{align}
    where
    \begin{equation*}
        \Sigma = \begin{pmatrix} \sigma_G^2 & 0 \\ 0 & \sigma_L^2 \end{pmatrix}.
    \end{equation*}
    Note that we can alternately solve for $\B{w}$ and $\sigma_G, \sigma_L$ as follows. First fix $\sigma_G\!=\!\sigma_L\!=\!1$ and compute
    \begin{equation*}
        \tilde{\B{w}}^{\dagger} = \left(\sum_{i=1}^n \Phi_i^{\top} \Phi_i + \lambda\, \B{I} \right)^{-1} \left(\sum_{i=1}^n \Phi_i^{\top} \B{x}_i\right) .
    \end{equation*}
    The optimal standard deviations $\sigma_G, \sigma_L$ are then simply given by the empirical standard deviations of the residuals under $\tilde{\B{w}}^{\dagger}$.
    Thus, the optimum of eq.~\eqref{eq:modela_optimization} is
    \begin{equation*}
        \B{w}^{\dagger} = \left(\sum_{i=1}^n \Phi_i^{\top} \Sigma^{-1} \Phi_i + \lambda\, \B{I} \right)^{-1} \left(\sum_{i=1}^n \Phi_i^{\top} \Sigma^{-1} \B{x}_i\right) ,
    \end{equation*}
    where $\Sigma = \diag(\sigma_G^2, \sigma_L^2)$.
    \item Given fitted weights $\B{w}^{\dagger}$, estimate the errors $\epsilon_G$, $\epsilon_L$,
    \begin{equation*}
        \hat{\B{\epsilon}}_i \equiv (\hat{\epsilon}_{g_i}, \hat{\epsilon}_{l_i})^{\top} \equiv \B{x}_i - \Phi_i \B{w}^{\dagger} .
    \end{equation*}
    \item Fit a counterfactually fair predictor $\hat y_i \equiv f_{\B{\theta}}(\hat{\B{\epsilon}_i})$ with parameters $\B{\theta}$ to predict outcomes $y_i$ via
    \begin{equation*}
        \B{\theta}^{\dagger} = \argmin_{\B{\theta}} \sum_{i=1}^n \mathcal{L}(f_{\B{\theta}}(\hat{\B{\epsilon}}_i), y_i) ,
    \end{equation*}
    for some loss function $\mathcal{L}$.
    While virtually any predictive model can be used in the two-variable case, in the general case we require the counterfactually fair predictor to be differentiable, such that it is amenable to gradient-based optimization.
    The definition of counterfactual fairness constrains the optimization for any loss function.
    Here, we use the sufficient condition for counterfactual fairness that the predictor $\hat{y}$ depends only on the error terms, which are non-descendants of $A$~\citep{kusner:17}.
\end{enumerate}

\subsection{MODEL B: EVALUATE CFU}

Next, we evaluate how the predictor $f_{\B{\theta}^{\dagger}}$ obtained in the previous section breaks down in the presence of unobserved confounding, i.e., in Model B.
To do so, we fit Model B and generate ``true'' counterfactuals $\B{x}'$.
If we were handed these counterfactuals and we wanted to make predictions using $f_{\B{\theta}^{\dagger}}$ we would compute their error terms $\hat{\B{\epsilon}}'$ using Step 2 above.
If Model A was in fact the model that generated the counterfactuals $\B{x}'$ then the predictions on the error terms for the real data and the counterfactuals would be \emph{identical}: $f_{\B{\theta}^{\dagger}}(\hat{\B{\epsilon}})\!=\!f_{\B{\theta}^{\dagger}}(\hat{\B{\epsilon}}')$.

However, because the counterfactuals were generated by the true weights $\B{w}^{*}$ of Model B, not the weights $\B{w}^{\dagger}$ of Model A, there will be a difference between the real data and counterfactual predictions $f_{\B{\theta}^{\dagger}}(\hat{\B{\epsilon}})\!\neq\!f_{\B{\theta}^{\dagger}}(\hat{\B{\epsilon}}')$.
It is this discrepancy we will quantify with our measure of counterfactual unfairness (CFU).
Here is how we compute it for a given confounding budget $p_{\max}$.
\begin{enumerate}[label=\textbf{\arabic*.},wide,labelwidth=!,labelindent=0pt]
    \item Fit model B via regularized maximum likelihood:
    \begin{align}\label{eq:modelb_optimization}
    \begin{split}
        \min_{\B{w}, \sigma_G, \sigma_L}
        &\sum_{i=1}^{n}
        (\B{x}_i - \Phi_i \B{w})^{\top} \Sigma^{-1} (\B{x}_i - \Phi_i \B{w}) \\
        &+ \lambda^{\dagger} \|\B{w} \|_2^2 + n\,\log \det(\Sigma) ,
    \end{split}
    \end{align}
    where
    \begin{align*}
        \Sigma &\equiv \begin{pmatrix} \sigma_G & 0 \\ 0 & \sigma_L \end{pmatrix}
        \underbrace{\begin{pmatrix} 1 & p_{\max} \\ p_{\max} & 1 \end{pmatrix}}_{P}
        \begin{pmatrix} \sigma_G & 0 \\ 0 & \sigma_L \end{pmatrix} .
    \end{align*}
    As before we can alternately solve for $\B{w}$ (closed-form) and $\sigma_G, \sigma_L$ (via coordinate descent).\footnote{In fact we optimize $\log(\sigma_G), \log(\sigma_L)$ to ensure the standard deviations are positive.} Let $\B{w}^*$ be the final weights after optimization.
    \item Given weights $\B{w}^*$, estimate the errors of Model B,
    \begin{equation*}
        \hat{\B{\delta}}_i = (\hat{\delta}_{g_i}, \hat{\delta}_{l_i})^{\top}
        = \B{x}_i - \Phi_i \B{w}^* .
    \end{equation*}
    \item For a fixed counterfactual value $a' \in \mathcal{A}$, compute the Model B counterfactuals of $G$ and $L$ for all $i$,
    \begin{align*}
        g'_i &= \B{\phi}_{G}(a'_i)^{\top} \B{w}^*_{G} + \hat{\delta}_{g_i}\;,\\
        l'_i &= \B{\phi}_{L}(a'_i, g'_i)^{\top} \B{w}^*_{L} + \hat{\delta}_{l_i} ,
    \end{align*}
    where $\B{w}^* = (\B{w}^*_{G}, \B{w}^*_{L})^{\top}$.
    If $\B{x}'_i \equiv (g_i', l_i')^{\top}$, we can write the above equation as
    \begin{equation*}
       \B{x}'_i = \Phi'_i \B{w}^* + \hat{\B{\delta}}_i ,
    \end{equation*}
    and $\Phi'_i \equiv \mbox{diag}(\B{\phi}_{G}(a'_i), \B{\phi}_{L}(a'_i, g'_i))$ is defined in general by sequential propagation of counterfactual values according to the ancestral ordering of the SCM.
    \item Compute the (incorrect) error terms of the counterfactuals using the same procedure as in step 2 of Section~\ref{subsec:model_a}, using weights $\B{w}^{\dagger}$ of Model A:
    \begin{equation*}
        \hat{\B{\epsilon}}'_i = (\hat{\epsilon}_{g'_i}, \hat{\epsilon}_{l'_i})^{\top} = \B{x}'_i - \Phi'_i \B{w}^{\dagger} .
    \end{equation*}
    Again, the predictions on the above quantity $f_{\theta^{\dagger}}(\hat{\B{\epsilon}}'_i)$ will differ from those made on the real-data error terms $f_{\theta^{\dagger}}(\hat{\B{\epsilon}}_i)$ (unless the counterfactuals were also generated according to model A).
    \item To measure the discrepancy, we propose to quantify counterfactual unfairness as the squared difference between the above two quantities:
    \begin{equation*}
        \CFU_i = (
        f_{\B{\theta}^{\dagger}}(\hat{\B{\epsilon}}_i) -
        f_{\B{\theta}^{\dagger}}(\hat{\B{\epsilon}}'_i)
        )^2 .
    \end{equation*}
    Ultimately, to summarize the aggregate unfairness, we will compute the average counterfactual unfairness:
    \begin{align}
        \CFU = \frac{1}{n} \sum_{i=1}^n \CFU_i \label{eq:cfu}.
    \end{align}
\end{enumerate}
A quick note: in the two-variable setting, given a confounding budget $p_{\max}$, the worst-case CFU occurs precisely at $p_{\max}$ (which need not be the case for multivariate confounding as we show in Appendix~\ref{sec:app:examples}).
Thus, the above procedure computes the maximum CFU with bivariate confounding budget equal to $p_{\max}$.
CFU measures how the counterfactual responses $\hat Y(a)$ and $\hat Y(a')$, defined using model A, differ ``in reality'', i.e., if model B is ``true''.
What qualifies as bad CFU is problem dependent and requires interaction with domain experts, who can make judgment calls about the plausibility of the misspecification $p_{\max}$ that is required to reach a breaking point.
Here, a breaking point could be the CFU of a predictor that completely ignores the causal graph.

To summarize: we learn $\hat Y \equiv f_{\B{\theta}^{\dagger}}$ as function of $X$ and $A$, where $X$ and $A$ are implicit in the expression of the (estimated) error terms $\B{\epsilon}$ that are computed using the assumptions of the working Model A.
We assess how ``unfair'' $\hat Y$ is by comparing for each data point the two counterfactual values $\hat Y(a) \equiv f_{\B{\theta}^{\dagger}}(\hat{\B{\epsilon}}_i)$ and $\hat Y(a') \equiv f_{\B{\theta}^{\dagger}}(\hat{\B{\epsilon}}'_i)$ where the ``true'' counterfactual is generated according to the world assumed by Model B.
The space of models to which Model B belongs is a continuum indexed by $p_{\max}$, which will allow us to visualize the sensitivity of Model A by a one-dimensional curve.
We will do this by finding the best fitting model (in terms of structural equation coefficients and error variances) at different values of $p_{\max}$, so that the corresponding CFU measure is determined by $p_{\max}$ only (results on the above law school model are shown in Section~\ref{sec:experiments}).
We assume that the free confounding parameter is not identifiable from data (as it would be the case if the model was linear and the edge $A \rightarrow L$ was missing, the standard instrumental variable scenario).

\section{TOOL \#2: OPTIMIZATION-BASED}
\label{sec:multivariate}

In this section, we generalize the procedure outlined for the two-variable case in Section~\ref{sec:bivariate} to the general case.

\subsection{NOTATION AND PROBLEM SETUP}

Besides the protected attribute $A$ and the target variable $Y$, let there be $m$ additional observed feature variables $X_j$ in the causal graph $\cG$ each of which comes with an unobserved error term variable $\epsilon_j$.

As before, we express the assignment of the structural equations for a specific realization of observed features $\B{x} = (x_1,\ldots,x_m)^{\top}$ and error terms $\B{\epsilon} = (\epsilon_1,\ldots,\epsilon_m)^{\top}$ as the following operation, i.e., $\B{x} = \Phi \B{w} + \B{\epsilon}$.
Here $\Phi$ has $m$ rows and $d = \sum_{V \in \text{has-parents}(\cG)} d_V$ columns, where $d_V$ is the dimensionality of embedding $\phi_{V}:\bR^{|\pa_{\cG}(V)|} \to \bR^{d_V}$ for each node $V \in \cG$ that has parent nodes.
Without loss of generality, we assume the nodes $\{A\} \cup \{X_j\}_{j=1}^m$ to be topologically sorted with respect to $\cG$ with $A$ always being first.
We combine the individual weights as,
$\B{w} = (\B{w}_{X_1}, \B{w}_{X_2}, \ldots, \B{w}_{X_m}) \in \bR^d$,
and represent $\Phi$ once evaluated on a specific sample $(a, \B{x})$ of the variables $A, X_1,\ldots,X_m$ as,
\begin{equation*}
    \Phi =
    \begin{pmatrix}
        \B{\phi}_{X_1}^{\top} &  & \B{0}^{\top}\\
        & \ddots & \\
        \B{0}^{\top} & & \B{\phi}_{X_m}^{\top}
    \end{pmatrix},
\end{equation*}
where $\B{\phi}_{X_j}$ is based on the parents of $X_j$, $(a,\B{x}_{\pa_G(X_j)})$.
The covariance matrix of the error terms is given by
\begin{equation*}
\Sigma \equiv \diag(\sigma_1, \dots, \sigma_m)P\diag(\sigma_1, \dots, \sigma_m),
\end{equation*}
\noindent where $\sigma_1, \dots, \sigma_m$ are the standard deviations of each variable and $P$ is a correlation matrix.

\subsection{THE OPTIMIZATION PROBLEM}

In the general case our goal is to find a correlation matrix $P$ that satisfies a ``confounding budget'' $p_{\max}$.
In particular we would like to constrain the correlation $P_{jk}$ between any two different variables $X_j$ and $X_k$ for $j \neq k$, while allowing $P_{jj} = 1$ for all $j$.
Additionally, we want to take into account any prior knowledge that certain variable pairs should have no correlation, if available.
The most intuitive way to budget the amount of confounding is to limit the absolute size of any correlation by $p_{\max}$ as: $|P_{jk}| \leq p_{\max}$ for all $j \ne k$.
This captures a notion of ``restricted unobserved confounding'' and leads
to the following optimization problem
\begin{align}
    \max_P &\quad\sum_{i=1}^n \CFU_i \label{eq:multiopt_orig} \\
    \text{subject to}
    &\quad P_{jj} = 1 \quad\text{for } j \in \{1, \ldots, m\}, \nonumber \\
    &\quad |P_{jk}| \le p_{\max} < 1 \quad\text{for all } (j, k)
      \in \mathcal{C}, \nonumber \\
    &\quad P_{jk}= 0 \quad\text{for all } (j, k).
      \not\in \mathcal{C}, j \neq k.   \nonumber
\end{align}
$\mathcal{C}$ is the set of correlations that should be non-zero.
A quick aside: the setting  where there are zero correlations can be captured using the standard acyclic directed mixed graph notation \citep[ADMG,][]{richardson:03}.
Specifically, this can be represented by ADMGs by removing bidirected edges between any two error terms whose correlation is fixed to zero.

As in the bivariate case, CFU is a direct function of $P$ only: all other parameters will be determined given the choice of correlation matrix by maximizing likelihood.
Note that eq.~\eqref{eq:multiopt_orig} contains multiple nested optimization problems (for the counterfactually fair model weights $\B{\theta}^{\dagger}$, and the weights and standard deviations of Models A and B).
To solve it efficiently, we will parameterize $P$ in a way that facilitates optimization via off-the-shelf, unconstrained, automatic differentiation tools.
We provide more details about computational bottlenecks in Appendix~\ref{sec:app:computational}.

\subsection{ALGORITHM}

We use the following approach to accommodate the constraints in eq.~\eqref{eq:multiopt_orig} in a way such that our algorithm does not require a constrained optimization subroutine.
Assume first that $P$ has no correlations that should be zero.
We compute $L L^{\top}$ for a matrix $L \in \bR^{m \times m}$, whose entries are the parameters we eventually optimize.
To constrain the off-diagonals to a given range and ensure that $P$ has $1$s on the diagonal, we define $P$ as,
\begin{equation*}
    P := \tanh_{p_{\max}}(L L^{\top}) := \B{I} + p_{\max} \, (\B{J} - \B{I}) \odot \tanh(L L^{\top}),
\end{equation*}
where $\odot$ denotes element-wise multiplication of matrices and $\B{J}$ is a matrix of all ones.
This way $P$ is symmetric, differentiable w.r.t.\ the entries of $L$, has $1$s on the diagonal, and its off-diagonal values are squashed to lie within $(-p_{\max}, p_{\max})$.
While it is natural to directly mask and clamp the diagonal, there are various ways to squash the off-diagonals to a fixed range in a smooth way, which bears close resemblance to barrier methods in optimization.
We choose $\tanh()$ because of its abundance in ML literature, but other forms of $P$ may work better for specific applications.
Note that this formulation does not guarantee $P$ to be positive-semidefinite.

In Algorithm~\ref{algo:mult_cfu}, we describe our procedure to maximize counterfactual unfairness given a confounding budget $p_{\max}$ and observational data $\{\B{x}_i, y_i, a_i, \}_{i=1}^n \subset \bR^m \times \{0,1\}^2$.
The algorithm closely follows the procedure described in Section~\ref{sec:bivariate} for the bivariate case.
Since we use automatic differentiation provided by PyTorch~\citep{paszke2017automatic} to obtain gradients, we only show the forward pass in Algorithm~\ref{algo:mult_cfu}.
For the initialization \textsc{InitializeParameters()}, we simply populate $L$ as a lower triangular matrix with small random values for the off-diagonals and $1$s on the diagonal.

If $\mathcal E$ indicates some correlations should be zero, we suggest the following standard ``clique parameterization'': $L$ is a $m \times c$ matrix where $c$ is the number of cliques in $\mathcal E $, with $L_{ik}$ being a non-zero parameter if and only if vertex $i$ is in clique $k$.
$L_{ik} \equiv 0$ otherwise.
It follows that such a matrix will have zeros at precisely the locations not in $\mathcal E$.\footnote{Barring unstable parameter cancellations that have measure zero under continuous measures on $\{L_{ik}\}$.}
See \cite{silva_nips:07} and \cite{barber:09} for examples of applications of this idea.
For large cliques, further refinements are necessary to avoid unnecessary constraints, such as creating more than one row per clique of size four or larger.
In the interest of space, details are left for an expanded version of this paper and our experiments will not make use of sparse $P$ (note that this parameterization also assumes that the number of cliques is tractable).
Note that individual parameters $L_{ik}$ may not be identifiable, but identifiability is not necessary here, all we care about is the objective function: CFU.
As a matter of fact, multiple globally optimal solutions are to be expected even in the space of $P$ transformations.
A more direct parameterization of sparse $P$, with exactly one parameter per non-zero entry of the upper covariance matrix, is discussed by \cite{drton:04}.
Computationally, this minimal parameterization does not easily lead to unconstrained gradient-based algorithms for optimizing sparse correlation matrices with bounded entries.
We suggest the clique parameterization as a pragmatic alternative.
Special cases may be treated with more efficient specialized approaches.
See \citep{cinelli19a} for a thorough discussion of fully linear models.

In Section~\ref{sec:experiments} we will demonstrate this approach on a 3-variable-confounding scenario to showcase our approach.
As this paper is aimed at describing the methodology, we will leave more complex confounding scenarios to an extended version of this work and only briefly describe an extension to path-specific effects in Appendix~\ref{sec:app:paths}.

\begin{algorithm}[t!]
\caption{\textsc{MaxCFU}: Maximize counterfactual unfairness under a certain confounding budget constraint.}\label{algo:mult_cfu}
\begin{algorithmic}[1]
    \Require
    data $\{\B{x}_i, y_i, a_i, \}_{i=1}^n$,
    confounding budget $p_{\max}$, learning rate $\alpha$, minibatch size $B$
    \Statex
  \State $\{\hat{\B{\epsilon}}_i\}_{i=1}^n, \B{w}^{\dagger}, \B{\theta}^{\dagger} \gets \Call{FitModelA}{\{\B{x}_i, y_i, a_i, \}_{i=1}^n}$
  \State $\cD \gets \{\B{x}_i, y_i, a_i, \Phi_i, \hat{\B{\epsilon}}_i \}_{i=1}^n$\Comment{full dataset}
  \State $L \gets \Call{InitializeParameters}{{}}$
  \For{$t = 1 \ldots T$}\Comment{iterations}
  \State $\cD^{(t)} \gets \Call{SampleMinibatch}{\cD, B}$
  \State $\Delta \gets \nabla_L \Call{CFU}{\cD^{(t)}, \B{w}^{\dagger}, \frac{B}{n}\lambda^{\dagger}, \B{\theta}^{\dagger}, L}$ \Comment{autodiff}
  \State $L \gets L + \alpha \Delta$ \Comment{gradient ascent step}
  \EndFor
  \State \Return $\Call{CFU}{\cD, \B{w}^{\dagger}, \lambda^{\dagger}, \B{\theta}^{\dagger}, L}$
  \Statex
  \Function{FitModelA}{$\{\B{x}_i, y_i, a_i, \}_{i=1}^n$}
      \State $\B{w}^{+} \gets \Big(\sum_{i=1}^n \Phi_i^{\top} \Phi_i + \lambda^{\dagger} \B{I} \Big)^{-1} \Big(\sum_{i=1}^n \Phi_i^{\top} \B{x}_i\Big)$
      \State $\Sigma \gets \diag(\mathrm{var}(\{\B{x}_i - \Phi_i \B{w}^{+}\}_{i=1}^n))$
      \State $\B{w}^{\dagger} \gets \Bigl(\sum_{i=1}^n \Phi_i^{\top} \Sigma^{-1} \Phi_i + \lambda^{\dagger} \B{I} \Bigr)^{-1} \times$\newline
      \hspace*{1.5cm}$\Bigl(\sum_{i=1}^n \Phi_i^{\top} \Sigma^{-1} \B{x}_i\Bigr)$
      \State $\hat{\B{\epsilon}}_i \gets \B{x}_i - \Phi_i \B{w}^{\dagger}$
      \State $\B{\theta}^{\dagger} \gets \argmin_{\B{\theta}} \sum_{i=1}^n \mathcal{L}(f_{\B{\theta}}(\hat{\B{\epsilon}}_i), y_i)$
      \State \Return $\{\hat{\B{\epsilon}}_i\}_{i=1}^n, \B{w}^{\dagger}, \B{\theta}^{\dagger}$
  \EndFunction
  \Statex
  \Function{CFU}{{$\cD$, $\B{w}^{\dagger}$, $\lambda^{\dagger}$, $\B{\theta}^{\dagger}$, $L$}}
    \State $\B{w}^*, \B{\sigma}^* \gets
      \min_{\B{w}, \B{\sigma}}
      \sum_{i=1}^{n}
      (\B{x}_i - \Phi_i \B{w})^{\top} \Sigma^{-1} \times$\newline
      \hspace*{2.2cm}$(\B{x}_i - \Phi_i \B{w})
      + \lambda^{\dagger} \|\B{w} \|_2^2 + n\,\log \det(\Sigma)$\newline
      \hspace*{0.5cm}where
      $\Sigma = \diag(\B{\sigma}) \tanh_{p_{\max}}(L L^{\top}) \diag(\B{\sigma})$
    \State $\hat{\B{\delta}}_i \gets \B{x}_i - \Phi_i \B{w}^*$
    \State $a'_i \gets 1 - a_i$ and $\B{x}'_i \gets \Phi'_i \B{w}^* +\hat{\B{\delta}}_i$\newline
      \hspace*{0.5cm}where $\Phi'_i$ is computed via iterative assignment
    \State $\hat{\B{\epsilon}}'_i \gets \B{x}'_i - \Phi'_i \B{w}^{\dagger}$
    \State $\CFU \gets \frac{1}{n} \sum_{i=1}^n (
          f_{\B{\theta}^{\dagger}}(\hat{\B{\epsilon}}_i) -
          f_{\B{\theta}^{\dagger}}(\hat{\B{\epsilon}}'_i)
          )^2$
    \State \Return $\CFU$
  \EndFunction
\end{algorithmic}
\end{algorithm}

\section{EXPERIMENTS}
\label{sec:experiments}

We compare the grid-based and the optimization-based tools introduced in Sections 3 and 4 on two real datasets.

In all experiments our embedding $\phi$ is a polynomial basis up to a fixed degree.
The degree is determined via cross validation (5-fold) jointly with the regularization parameter $\lambda^{\dagger}$.
Our counterfactually fair predictor is regularized linear regression on the noise terms $\B{\epsilon}$:
\begin{equation*}
    \min_{\B{\theta}} \sum_{i=1}^n (y_i - \B{\phi}(\hat{\B{\epsilon}}_i)^{\top} \B{\theta})^2 + \lambda \|\B{\theta} \|_2^2 \, .
\end{equation*}
For this model, counterfactual unfairness is:
\begin{equation*}
    \CFU_i = \Bigl(\bigl(\B{\phi}(\hat{\B{\epsilon}}_i) - \B{\phi}(\hat{\B{\epsilon}}'_i)\bigr)^{\top} \B{\theta}^{\dagger} \Bigr)^2 .
\end{equation*}

For comparison, we also train two baselines that also use regularized ridge regression (degree and regularization are again selected by 5-fold cross-validation):
\begin{description}[leftmargin=*,topsep=0pt,noitemsep]
    \item[unconstrained:] an unconstrained predictor using all observed variables as input $f_{\mathrm{uc}}: (A, X_1, \ldots, X_m) \mapsto Y$.
    \item[blind unconstrained:] an unconstrained predictor using all features, but not the protected attribute, as input $f_{\mathrm{buc}}:(X_1, \ldots, X_m) \mapsto Y$.
\end{description}
Analogous to our definition of $\CFU$ in eq.~\eqref{eq:cfu}, we compute the unfairness of these baselines as the mean squared difference between their predictions on the observed data and the predictions of the counterfactually fair predictor on the observed data: $\frac{1}{n}\sum_{i=1}^n (f_{\B{\theta}^{\dagger}}(\hat{\B{\epsilon}}_i) - \hat{y}_i^{\mathrm{(b)uc}})^2$, where $\hat{y}_i^{\mathrm{uc}} = f_{\mathrm{uc}}(a_i, \B{x}_i)$ and $\hat{y}_i^{\mathrm{buc}} = f_{\mathrm{buc}}(\B{x}_i)$.
This choice is motivated by the fact that in practice we care about how much potential predictions deviate from predictions satisfying a fairness measure. For our grid-based approach we repeatedly fix $p_{\max} \in [0, 1)$ to a particular value and then use the procedure in Section~\ref{sec:bivariate} to compute $\CFU$.
For the optimization approach we similarly fix $p_{\max} \in [0, 1)$ in the constraint of eq.~\eqref{eq:multiopt_orig}.
For efficiency we use the previously found correlation matrix $P$ as initialization for the next setting of $p_{\max}$.

\begin{figure}
    \centering
    \hspace{1cm}\textbf{Law School}\\
    \includegraphics[width=0.9\columnwidth]{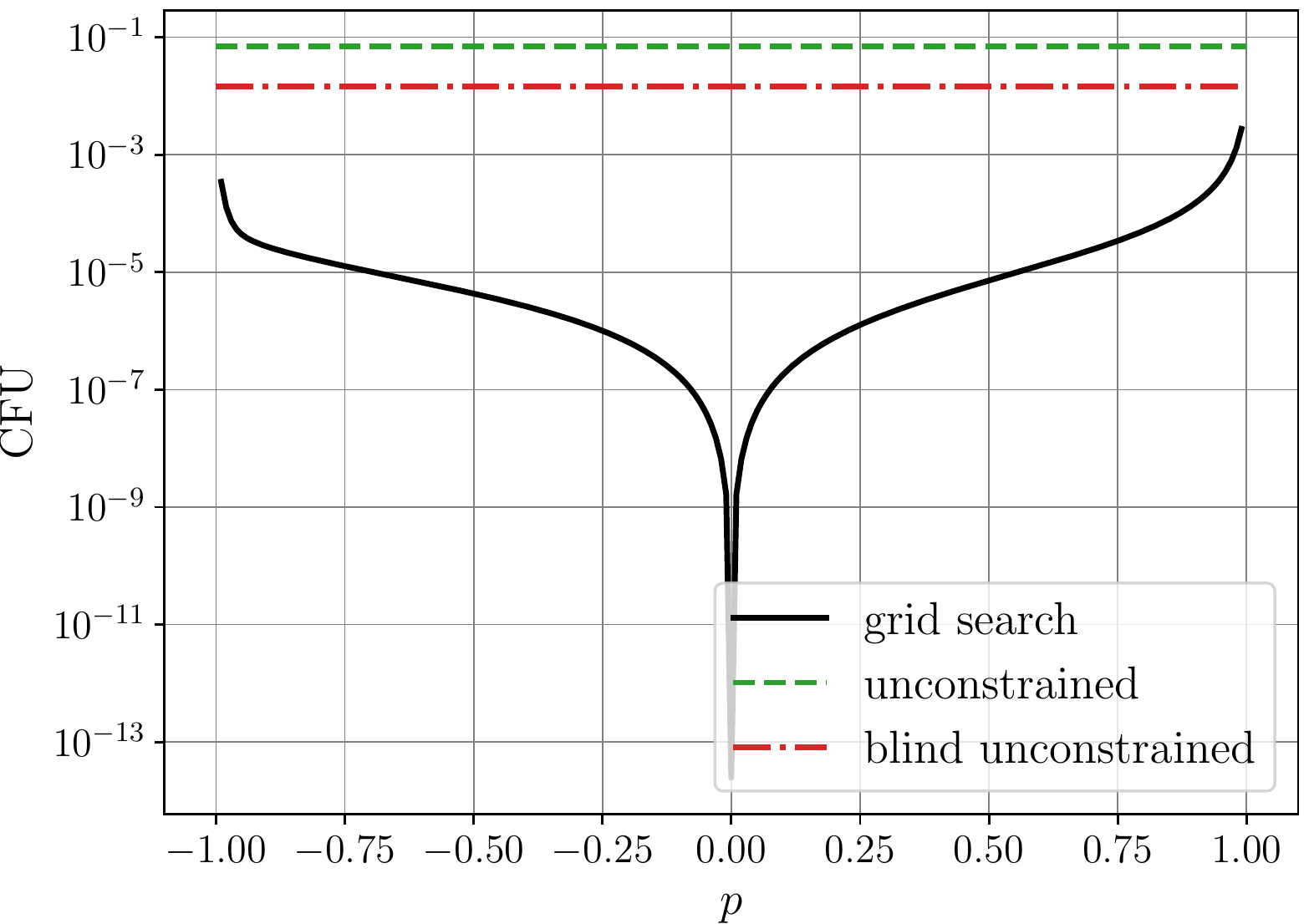}
    \caption{Counterfactual unfairness for the law school dataset. See text for details.}
    \label{fig:results_bivariate}
\end{figure}

\begin{figure}[t]
    \centering
    \begin{tikzpicture}
    \node[protected, label=right:{\color{Orange}race}] (A) {$A$};
    \node[latent, left=of A] (EO) {$\epsilon_O$};
    \node[latent, right=of A] (EM) {$\epsilon_M$};
    \node[feature, below left=0.5 of A] (O) {$O$};
    \node[feature, right=of O] (M) {$M$};
    \node[feature, below right=0.5 of O] (J) {$J$};
    \node[latent, right=of J] (EJ) {$\epsilon_J$};
    \edge {A} {O,M,J};
    \edge {O,EM} {M};
    \edge {O,M} {J};
    \edge {EO} {O};
    \edge {EJ} {J};
    \draw[dashed, <->] (EO) to[bend left] (EM);
    \draw[dashed, <->] (EO) to[out=35,in=160] ($(EM) + (0.1,0.5)$) to[out=340,in=70] (EJ);
    \draw[dashed, <->] (EM) to (EJ);
    \node[left=1.0 of O, ForestGreen, align=left] (label) {%
    \textbf{features}\\[3pt]
    satisfaction with\\[3pt]
    $O$: organization\\[2pt]
    $M$: manager\\[2pt]
    $J$: job};
    \end{tikzpicture}%
    \caption{The true causal graph (Model B) for the NHS Staff Survey dataset.}
    \label{fig:nhs}
\end{figure}
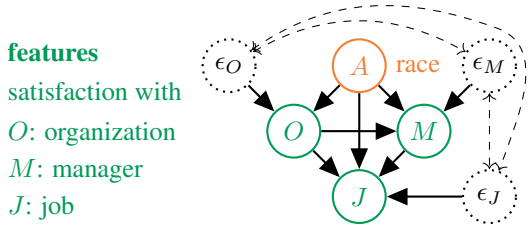

\begin{figure}
    \centering
    \hspace{1cm}\textbf{NHS}\\
    \includegraphics[width=0.9\columnwidth]{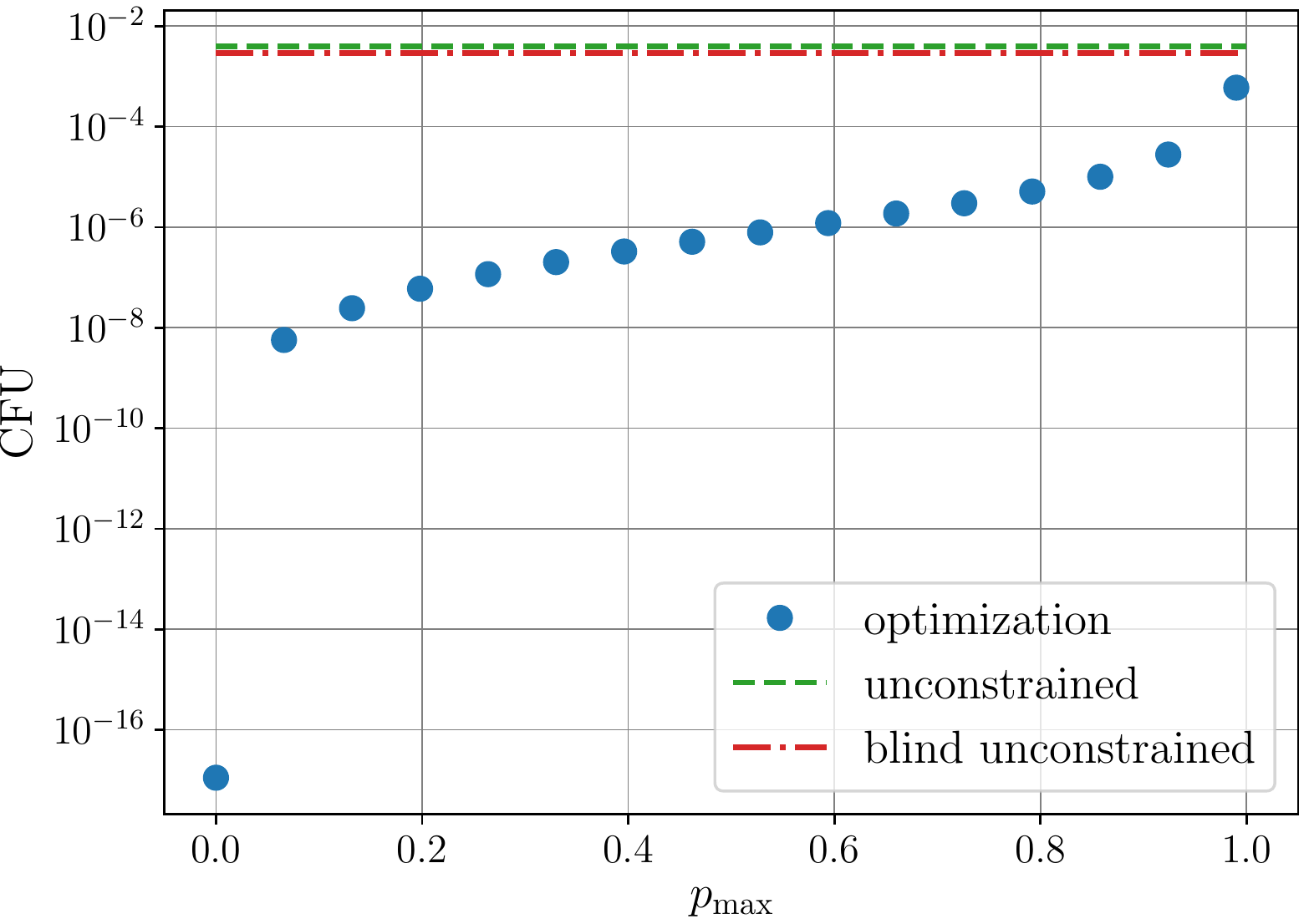}
    \caption{Counterfactual unfairness as a function of $p_{\max}$ for the multivariate NHS dataset.}
    \label{fig:results_multivariate}
\end{figure}

\paragraph{Law School data.}
Our first experiment is on our motivating example introduced in Section~\ref{sec:bivariate} on law school success (recall eq.~\eqref{eq:law} and Figure~\ref{fig:bivariate} for details on the causal models). Our data comes from law school students in the US \citep{wightman1998lsac}. As our causal model investigates confounding between two variables we will use the grid-based approach introduced in Section~\ref{sec:bivariate} to calculate the maximum $\CFU$. Recall that for the bivariate approach we fix a confounding level $p\!=\!p_{\max}$ and then compare predictions between real data based on Model A versus counterfactuals generated from Model B.  Figure~\ref{fig:results_bivariate} shows the $\CFU$ for the grid-based approach (black), alongside the baselines (green/red), as the correlation $p$ varies. We first note that the confounding is not symmetric around $p\!=\!0$. For the law school data, negative correlations have smaller $\CFU$. In general, this is a data-specific property.

Additionally, we notice that as $p_{\max}$ moves away from $0$ it increases noticeably, then plateaus in roughly $[0.1, 0.9]$ and finally increases again. Our suspicion is that the initial jump may be due to a small model misspecification. Specifically, a small change in $p_{\max}$ may cause the generated counterfactuals to have additional error which may dominate for such small $p_{\max} \leq 0.1$ and then becomes insignificant for larger $p_{\max}$. For large $p_{\max}\geq 0.9$ we believe the increase may be due to numeric instability as the covariance matrix becomes nearly negative definite. This could cause the weights of the model to rapidly grow or shrink.
Between the small and large regimes, we see the $\CFU$ gradually increase as $p_{\max}$ is increased. Finally, we note that both baseline approaches have higher $\CFU$ than found with any grid-based setting.

\paragraph{NHS Staff Survey.}
Our second experiment is based on the 2014 UK National Health Service (NHS) Survey \cite{nhs2014staffsurvey}. The goal of the survey was to ``gather information that will help to improve the working lives of staff in the NHS''. Answers to the survey questions ranged from `strongly disagree' (1) to `strongly agree' (5). We averaged survey answers for related questions to create a dataset of continuous indices for: \emph{job satisfaction} $(J)$, \emph{manager satisfaction} $(M)$, \emph{organization satisfaction} $(O)$, and \emph{overall health} $(Y)$. The goal is to predict health $Y$ based on the remaining information. Additionally, we collected the race $(A)$ of the survey respondents. Using this data, we formulate a ground-truth causal graph shown in Figure~\ref{fig:nhs} (equivalent to Model B in Figure~\ref{fig:bivariate}). This causal graph includes correlations between all error terms $\epsilon_J, \epsilon_M, \epsilon_O$. This model has the following structural equations
\begin{align}
    O &= \B{\phi}_O(A)^{\top} \B{w}_{O} + \epsilon_O  \label{eq:nhs} \\
    M &= \B{\phi}_M(A,O)^{\top} \B{w}_{M} + \epsilon_M  \nonumber \\
    J &= \B{\phi}_J(A,O,M)^{\top} \B{w}_{J} + \epsilon_J . \nonumber
\end{align}

Just as in the law school example, we measure the impact of this confounding by comparing this model to the unconfounded model (i.e., all error terms are jointly independent). As there is no general efficient way to grid-search for positive definite matrices that maximize $\CFU$ for a given $p_{\max}$, we make use of our optimization-based procedure for calculating maximum $\CFU$, as described in Algorithm~\ref{algo:mult_cfu}.  Figure~\ref{fig:results_multivariate} shows the results of our method on the NHS dataset. Note that we only show positive $p_{\max}$ because our optimization problem eq.~\eqref{eq:multiopt_orig} only constrains the absolute value of the off-diagonal correlations. This allows the procedure to learn whether positive or negative correlations result in greater $\CFU$. As in the law school dataset we see an initial increase in CFU for small $p_{\max}$, followed by a plateau, ending with another small increase. As before, all values have lower values than the two baseline techniques.

\section{CONCLUSION}
\label{sec:conclusion}
In this work we presented two techniques to assess the impact of unmeasured confounding in causal additive noise models. We formulated unmeasured confounding as covariance between error terms. We then introduced a grid-based approach for confounding between two terms, and an optimization-based approach for confounding in the general case. We demonstrated our approach on two real-world fairness datasets. As a next step, we plan to write an extended version of this work with experiments on larger graphs with known zero correlations. We would also like to extend these approaches to handle the sensitivity of other quantities such as the structural equations.  Overall, we believe the tools in this work are an important step towards making causal models suitable to address discrimination in real-world prediction problems.

\section*{Acknowledgments}
We thank Chris Russell for useful discussions during the initial phase of this project.
AW, MK, RS acknowledge support from the David MacKay Newton research fellowship at Darwin College, The Alan Turing Institute under EPSRC grant EP/N510129/1 \& TU/B/000074, and the Leverhulme Trust via the CFI.

\bibliography{bibliography}
\bibliographystyle{icml2019}

\clearpage
\appendix
\section{COUNTEREXAMPLE}
\label{sec:app:examples}

In this section we show that in the multivariate setting, the worst-case counterfactual unfairness with a confounding budget of $p_{\max}$ is not necessarily obtained when all non-zero entries of the correlation matrix are set to $p_{\max}$.
To this end, it suffices to find a symmetric matrix $A$ with $1$s on the diagonal that is not positive-semidefinite when all its non-zero off-diagonal entries are set to the same value, which we define to be the considered confounding budget $p_{\max}$.
Since each valid correlation matrix must be positive-semidefinite, the correlation matrix for the worst-case counterfactual unfairness must be different from $A$ (while maintaining the zero entries).
Because all off-diagonal entries are upper bounded by $p_{\max}$, at least one of them must be smaller than the corresponding value in $A$.

For example, consider
\begin{equation*}
A = \begin{pmatrix}
1 & p_{\max} & p_{\max} \\
p_{\max} & 1 & 0 \\
p_{\max} & 0 & 1
\end{pmatrix}.
\end{equation*}
Since the eigenvalues of $A$ are $1$, $1 - \sqrt{2} p_{\max}$, and $1 + \sqrt{2} p_{\max}$, we see that $A$ is not positive-semidefinite for $p_{\max} > 1/\sqrt{2}$.

In general, the matrix $A \in \bR^{n \times n}$ with $A_{ii} = 1$ for $i \in \{1, \ldots, n \}$, $A_{1i} = A_{i1} = p_{\max}$ for $i \in \{2, \ldots, n\}$ and $A_{ij} = 0$ for all remaining entries, has the eigenvalues (without multiplicity) $1$, $1 - \sqrt{n-1} p$, and $1 + \sqrt{n-1} p$.
Therefore, $A$ is not positive-semidefinite for $p_{\max} > 1 / \sqrt{n-1}$.
We conclude that as the dimensionality of the problem increases, we may encounter such situations for ever smaller confounding budget.

\section{COMPUTATIONAL CONSIDERATIONS}
\label{sec:app:computational}

Step 17 of Algorithm~\ref{algo:mult_cfu} is the main place where code optimization can take place, and alternatives to the (local) penalized maximum likelihood taking place there could be suggested (perhaps using spectral methods).
It is hard though to say much in general about Step 20, as counterfactual fairness allows for a large variety of loss functions usable in supervised learning.
In the case of linear predictors, it is still a non-convex problem due to the complex structure of the correlation matrix, and for now we leave as an open problem whether non-gradient based optimization may find better local minima.

\section{PATH-SPECIFIC SENSITIVITY}
\label{sec:app:paths}

Path-specific effects were not originally described by \cite{kusner:17} as the goal there was to introduce the core idea of counterfactual fairness in a way as accessible as possible (some discussion is provided in the supplementary material of that paper). See \cite{chiappa:18} for one take on the problem, and \cite{loftus2018causal} for another take to be fully developed in a future paper. Here we consider an example that illustrates how notions of path-specific effects \citep{shpitser:13} can be easily pipelined with our sensitivity analysis framework.

\begin{figure}[t]
    \centering
       \begin{tikzpicture}[thick]
    \node[protected] (A) {$A$};
    \node[latent, left=of A] (EG) {$\epsilon_U$};
    \node[latent, right=of A] (EL) {$\epsilon_F$};
    \node[unfair, below left=0.5 of A] (G) {$U$};
    \node[target, right=of G, label=right:{\color{white}features}] (L) {$F$};
  \edge {A} {G,L};
    \edge {EL} {L};
    \edge {EG} {G};
   \draw[dashed, <->] (EG) to[bend left] (EL);
    \end{tikzpicture}
    \caption{A path-specific model where the path from protected attribute $A$ to feature $U$ is unfair and the path from $A$ to feature $F$ is fair.}
    \label{fig:path}
\end{figure}
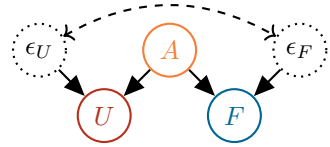

Consider Figure~\ref{fig:path}, where the path from $A \rightarrow U$ is considered unfair and $A \rightarrow F$ is considered fair, in the sense that we do not want a non-zero path-specific effect of $A$ on $\hat Y$ that is comprised by a possible path $A \rightarrow U \rightarrow \hat Y$ in the causal graph implied by the chosen construction of $\hat Y$. Then a path-specific counterfactually fair predictor is one that uses $\{\epsilon_U,F\}$ as input. Note that the only difference this makes in our grid-based tool is that we only estimate the error $\epsilon_U$ for the unfair path in Model A (step 2, Section 3.3) and fit a predictor on  $\{\epsilon_U,F\}$ (step 3, Section 3.3). Additionally, we only compute the incorrect error terms of the counterfactuals in Model B, using the weights of Model A (step 4, Section 3.4).  For the optimization-based tool we would change lines 13, 14, and 20 in the same way.

\end{document}